\documentclass[runningheads]{llncs}

\usepackage{eccv}

\usepackage{eccvabbrv}

\usepackage{graphicx}
\usepackage{booktabs}

\usepackage[accsupp]{axessibility}  

\usepackage{hyperref}

\usepackage{orcidlink}

\usepackage{nicefrac}
\usepackage{float}
\usepackage{subcaption}
\usepackage{multirow}
\usepackage{adjustbox}
\usepackage{algorithm}
\usepackage{algpseudocode}

\usepackage{soul}
\usepackage{marvosym}
\usepackage{amsfonts}
\usepackage{amsmath}
\usepackage{xcolor}
\usepackage{bbm}
\usepackage{siunitx}
\usepackage{lipsum}
\usepackage{nicefrac}

\definecolor{darkgreen}{rgb}{0.0, 0.7, 0.0}  

\DeclareMathOperator*{\argmin}{argmin} 

\begin{document}

\title{Vision Verification Enhanced Fusion of VLMs for Efficient Visual Reasoning} 

\titlerunning{V3Fusion of VLMs for Efficient Visual Reasoning}

\author{Selim Furkan Tekin\inst{1}\orcidlink{0000-0002-8662-3609} \and
Yichang Xu\inst{1}\orcidlink{0009-0000-1094-4206} \and
Gaowen Liu\inst{2}\orcidlink{0009-0000-9194-1233} \and Ramana Rao Kompella\inst{2}\orcidlink{0000-0002-7559-8997} \and Margaret L. Loper\inst{1}\orcidlink{0000-0002-0977-696X} \and Ling Liu\inst{1}\orcidlink{0000-0002-4138-3082} }

\authorrunning{Preprint}

\institute{Georgia Institute of Technology \and Cisco Systems \\
}
\maketitle


\begin{abstract}
With the growing number and diversity of Vision-Language Models (VLMs), many works explore language-based ensemble, collaboration, and routing techniques across multiple VLMs to improve multi-model reasoning. In contrast, we address the diverse model selection using both vision and language modalities. We introduce focal error diversity to capture complementary reasoning across VLMs and a CKA-based focal diversity metric (CKA-focal) to measure disagreement in their visual embeddings. On the constructed ensemble surface from a pool of candidate VLMs, we applied a Genetic Algorithm to effectively prune out those component VLMs that do not add value to the fusion performance. We identify the best combination for each task as well as fuse the outputs of each VLMs in the model pool, and show that heterogeneous models can capture epistemic uncertainty dynamically and mitigate hallucinations. Our V3Fusion approach is capable of producing dual focal-diversity fused predictions with high performance for vision-language reasoning, even when there is no majority consensus or the majority of VLMs make incorrect predictions. Extensive experiments validate V3Fusion on four popular VLM benchmarks (A-OKVQA, MMMU, MMMU-Pro, and OCR-VQA). The results show that V3Fusion outperforms the best-performing VLM on MMMU by 8.09\% and  MMMU-Pro by 4.87\% gain in accuracy. For generative tasks, V3Fusion outperforms Intern-VL2-8b and Qwen2.5-VL-7b, the top-2 VLM performers on both A-OKVQA and OCR-VQA. Our code and datasets are available at \url{https://github.com/sftekin/v3fusion}.
\end{abstract}    
\section{Introduction}
\label{sec:intro}

The modern Large Vision Language Models \cite{wu2024deepseek, Qwen2-VL, Qwen2.5-VL, liu2023visual, chen2024internvlscalingvisionfoundation, chen2024fargpt4vclosinggap} show remarkable performance on visual question answering and visual reasoning tasks. The models can match the information extracted from the visual with the input text and create reasoning steps, such as solving graduate-level multiple-choice questions \cite{yue2024mmmu} or generating commonsense reasoning on objects presented in the visual \cite{schwenk2022okvqa}.  These models are characterized by architectures with billions of parameters, massive training datasets, and remarkable performance across many zero and one-shot tasks. This results in their capabilities being diverse, making them experts in various subjects. Different from standard text-only LLMs, the VLMs are found on the visual extractor \cite{li2023blip2bootstrappinglanguageimagepretraining, li2022blipbootstrappinglanguageimagepretraining, zhai2023sigmoidlosslanguageimage, radford2021learningtransferablevisualmodels, jia2021scalingvisualvisionlanguagerepresentation}, the language decoder \cite{touvron2023llama, achiam2023gpt, team2024gemma, jiang2024mixtral, vaswani2017attention}, and their compatibility together. However, a VLM can extract the visual information thoroughly, but fails to make deep connections and reasoning based on the given query. Or it may have a deeper reading capacity but fail to detect or misidentify the objects in the image. Despite the existence of these flaws, a VLM can continue to generate an output for an object it didn't see or answer a question it doesn't know. i.e. \textit{hallucinate}.
In this paper, we argue that one can compare and contrast the capabilities of VLMs with each other to find the best model combination and generate the most refined and accurate output. Based on this comparison, we also demonstrate that one can model the total uncertainty and detect when the model hallucinate and be rectified its illusion. 

However, this task carries significant challenges. First, from a large pool of VLMs, one needs to select the best model combination for the ensemble set by analyzing their outputs. The number of combinations that can be produced grows exponentially with the number of candidate models. Second, each model in the set can produce conflicting outputs; therefore, the generation mechanism should establish the best result for the target task by resolving the conflicts. Third, the source of uncertainty could be inherent in the data or model parameters, and an uncertainty threshold should be dynamically determined to reject highly uncertain answers. Lastly, the rejected samples should be rectified to produce a better answer. Recent works approach the challenges by prompting an auxiliary LLM to generate an uncertainty score \cite{manakul2023selfcheckgpt, xia2025survey} or creating a debate environment \cite{zhao2025auto, liang2023encouraging} or target the language embeddings for heterogeneous LLMs \cite{huang2024ensemble}. However, these methods are prompt-dependent, costly, and they do not target modalities in VLMs. 


 To this end, we introduce \underline{V}isual \underline{V}erification \underline{V}LM \underline{Fusion}, \texttt{V3Fusion}, visual and language diversity-optimized VLM ensemble method with five unique properties: (i) a Focal Central Alignment Kernel (CKA), or \textit{Focal-CKA}, to measure the average similarity between visual encoder embeddings in an ensemble set, (ii) \textit{Focal-Diversity} metric to capture the error diversity and the diversity-performance correlation among component VLMs of an ensemble; (iii) a diversity-optimized ensemble pruning algorithm to identify and select the best sub-ensembles from a pool of $N$ base LLMs; 
 (iv) V3Fusion-MLP and V3Fusion-LED are more specific yet powerful fusion models for multiple-choice and open-ended questions. V3Fusion-MLP leverages the probability distribution assigned to choices by candidate models, learning to detect patterns among model logits and find the correct solution even when the majority of candidate models are incorrect. V3Fusion-LED, learns to generate answers for open-ended questions by comparing and contrasting base-model outputs. (v) Finally, we develop a reflection mechanism based on the epistemic uncertainty of the system to reject or accept the generated output based on an adaptive threshold algorithm, \textit{adaptive epistemic uncertainty}.

\section{Related Work}
\label{sec:related_work}


For the Ensemble Learning in LLMs, classical methods such as majority voting are employed to ensemble multiple outputs generated from the same LLM,  \cite{wang2022self, li2024more}. The confidence score calculated by either a model \cite{yao2024tree} or deduced from the model itself by the logits of tokens produced by the model is also used for voting \cite{zhang2024truthx, li2024inference}. Another thread of literature is creating a debate environment \cite{liang2023encouraging, wan2024knowledge, du2023improving, chan2023chateval} with the cost of lengthy and complex prompt strategies. In addition, several supervised summarization LLM ensemble methods are proposed, such as LLM-Blender \cite{jiang2023llm} and TOPLA-Summary \cite{tekin2024llm}. These methods formalize the ensemble as a summarization problem using a seq2seq model. Moreover, models with Mixture-of-Expert (MoE) layers are introduced \cite{shazeer2017outrageously, jiang2023mistral, jiang2024mixtral}, which are also used by the authors of \cite{wu2024deepseek} to create a VLM model and for alignment through fusion in \cite{tekin2024h}. Alternatively, a distillation strategy is proposed in \cite{wan2024knowledge} by performing a token alignment on the probability distributions of the models. Regarding the model routing, many works such as \cite{chen2023frugalgpt, zhang2025router, ong2024routellm} reduced the cost of inference by performing prompt adaptation, caching, and model tuning to choose the strongest model in the pool. LLM-based Multi-Agents \cite{guo2024large, du2023improving} carry similar motivations in terms of exploiting multiple LLMs to work collaboratively for a particular task. However, these works do not incorporate the visual modality while performing ensemble, routing, or collaboration.

In the domain of ensemble learning for VLMs, \cite{lu2023beyond} made logit level aggregation for zero-few shot learning, \cite{agnolucci2023eco} focused on the visual encoder and learn an ensemble of prompts for image classification, and \cite{chen2025cluster} ensembles prompts in VLM adaptation. Many works \cite{tran2025multi, guo2024large} employed ensemble learning in their multi-agent frameworks, such as video understanding \cite{chen2025lvagent} and propose the majority option as a pseudo-label and evaluates the agents. Authors of \cite{sivakumaran2025dart} introduced multi-VLM debate with external expert tools. Closest to our work, the authors of \cite{alazraki2023not} explore ensemble strategies for VQA models. However, the authors shows a high potential for ensemble, yet the classical methods applied on cumulative logits of generated tokens, e.g. majority voting, weighted voting, and training a classifier to learn the weights, cannot improve the best model performance. Thus, they use an external LLM (PaLM \cite{anil2023palm}) to evaluate the outputs based on the self-reflection of each candidate model. Differently, as we show in this paper, the probability assigned to tokens representing the choices is highly significant, and a trained ensemble model can improve the base model's performance significantly without the need for an external LLM.

\section{Preliminaries}
\label{sec:problem_definition}

\begin{figure}[hbt!]
    \centering
    \includegraphics[width=0.6\linewidth]{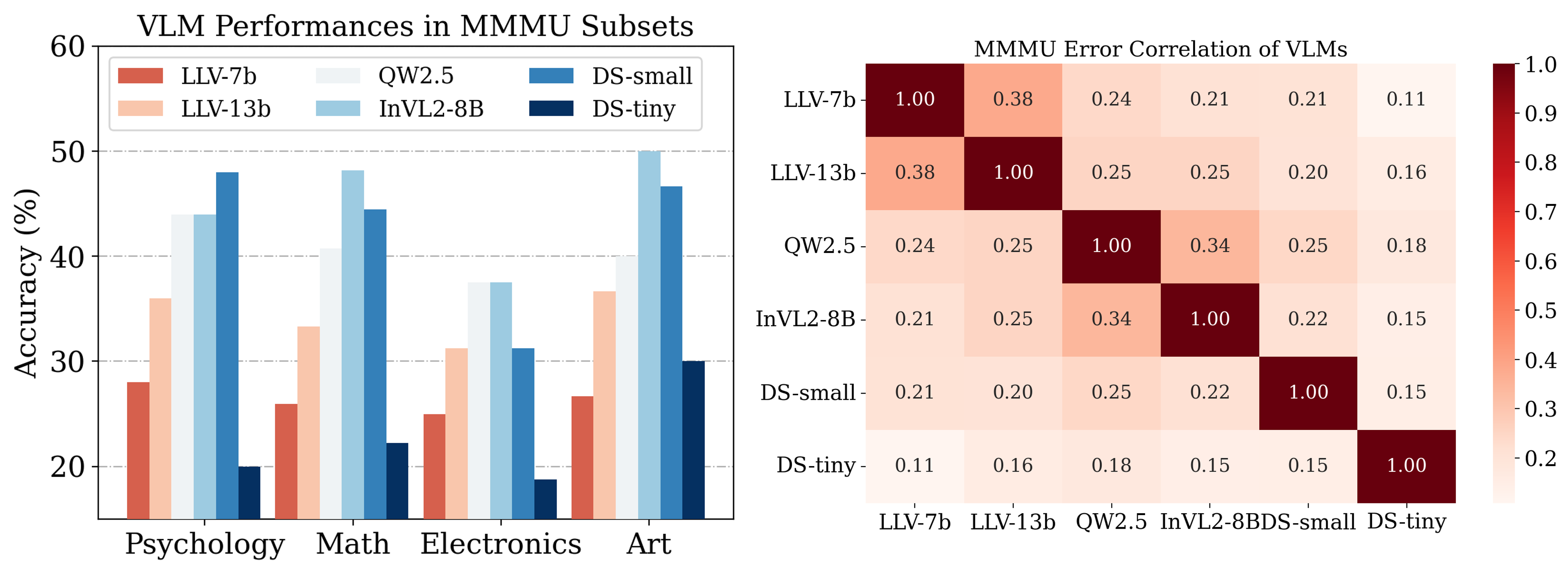}
    \caption{Performance of open-sourced and most popular VLMs in various tasks and their error correlation in MMMU.}
    \label{fig:motivation}
    \vspace{-8pt}
\end{figure}

\subsection{Diverse Capabilities of VLMs}

The Figure \ref{fig:motivation} shows an experimental proof on the diverse capabilities of VLMs, which caused by their structural differences such as, architectures, vocabularies, hidden sizes, training corpora and their instruction tuning strategies. As shown in the bar plot on the left of Figure \ref{fig:motivation}, the performances of different VLMs vary in different categories of college subjects. For example, while Deepseek-small is the best performing model in Psychology, it performs third-best in Electronics. A similar phenomenon can be observed for the other models, such as InternVL2 and Qwen. As our second insight, we observe that overall error correlation among these models is less. The ones with the high error correlations are within the models that have the same architectures, as shown in the correlation plot in Figure \ref{fig:motivation}, indicating redundancy. On contrary, the correlations between models that have different architectures are low, highlighting their importance in an ensemble pool. Based on this motivation, we aim for a mechanism that will select candidate models derived from their error correlations, while motivating their individual expertise.

\subsection{Multiple VLMs Ensemble Problem}

Let $(\mathbf{x}, \mathbf{I})$ denote an input tuple of a language instruction and an image, where the instruction $\mathbf{x}=\{\omega_1, \dots, \omega_L\}$ is composed of $L$ tokens $\omega_i$, and the image $\mathbf{I}\in\mathbb{R}^{H\times W\times 3}$ represents the RGB input. For task $T$ under a VLM $\mathcal{M}$ and let $\mathbf{y}$ denote the desired output. We assume a dataset $\mathcal{D}$ to be the collection of observations for task $T$, such that $(\mathbf{x}, \mathbf{I}, \mathbf{y}) \in \mathcal{D}$, where $|\mathcal{D}|=K$. For a pool of VLMs with the size $N$, denoted as $\mathcal{M}_1, \dots, \mathcal{M}_N$, we utilize $\mathcal{D}$ to find the optimal ensemble function $f_{\mathrm{ens}}$. This function takes outputs of each LLM and yields one final answer, denoted as $\widetilde{y}$, given by $f_{\mathrm{ens}}(\mathcal{M}_{1}(\mathbf{x}, \mathbf{I}), \dots, \mathcal{M}_{N}(\mathbf{x}, \mathbf{I}))=\widetilde{y}$, such that the difference between desired output is minimized, measured by the loss function $\mathcal{L}(\widetilde{y}, \mathbf{}y)$. However, based on the task $T$, the desired output $y$ can represent different solution spaces. For multiple-choice questions (MCQ), $y$ is a single token, $y^{(1)} \in \{1, \dots, m\}$ representing the choices where $m$ is number of choices. For open-ended questions (OEQ), $y$ consists of multiple tokens representing a single word, a real number, or a sequence of words and is denoted as $y^{(2)} = \{\omega_1, \dots, \omega_{L}\}$, where $L$ is the sequence length.

\subsection{Closer Look at Visual Language Models}
\label{sec:closer}

\begin{figure*}[t]
    \centering
    \includegraphics[width=1\textwidth]{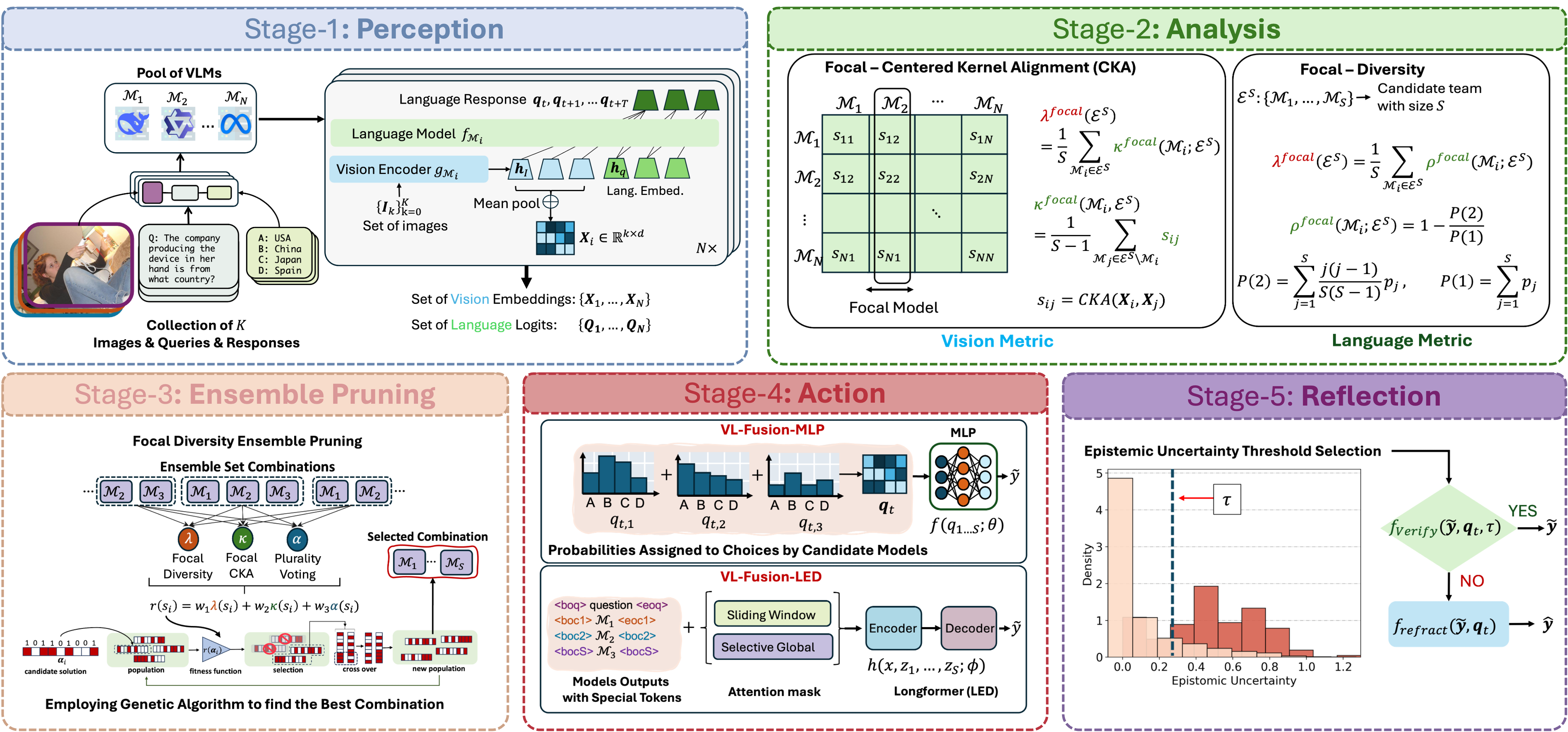}
    \vspace{-12pt}
    \caption{The V3Fusion Framework: An overview.}
    \label{fig:ensemble}
    \vspace{-10pt}
\end{figure*}

A standard VLM consists of two parts, visual encoder and language decoder as shown in the first step of Figure \ref{fig:ensemble}, \textit{Perception}. The visual encoder of model $\mathcal{M}_i$, denoted by $g_{\mathcal{M}_i}$, takes the an input image and produces visual embeddings $g_{\mathcal{M}_i}(\mathbf{I}_k)=[\mathbf{h}_{I,1}', \dots, \mathbf{h}_{I,L'}']$ and most VLMs e.g. \cite{liu2023visual} are using Vision Transformers (ViT) such as CLIP \cite{radford2021learning} and use a projection matrix to map the image embeddings to language space to generate visual token embeddings. For each token $\mathbf{h}_{I} = \mathbf{W}\cdot \mathbf{h}_{I}'$, where $\mathbf{h}_{I}\in\mathbb{R}^d$, model creates the visual embeddings $\mathbf{X}=[\mathbf{h}_{I,1},\dots, \mathbf{h}_{I,L'}]$. In the second phase, the language model takes language token embeddings $\mathbf{Q}=[\mathbf{h}_{q,1}, \dots,\mathbf{h}_{q,L}]$ together with the visual tokens to generate the first step of the response, $t$, denoted by $f_{\mathcal{M}_{i}}(\mathbf{X}, \mathbf{Q}) = \mathbf{q}_t$, where $\mathbf{q}_t\in\mathbb{R}^{d}$. 

The language model then, auto regressively predicts the next token, $\omega_{t+1}$,  by employing the \textit{softmax} function to model the probability mass conditioned on the input vision and language tokens $\mathbf{X}, \mathbf{Q}$, and the formerly generated tokens, $\omega_{<t}$, i.e. it models:
\vspace{-6pt}
\begin{equation}
\begin{split}
    p(y|\mathbf{I}, \mathbf{x}) = \prod_{t=1}^{L} p_{\mathcal{M}_i}&\big(y_t|y_{<t}, g_{\mathcal{M}_i}(\mathbf{I}),f_{\mathcal{M}_i}(\mathbf{x})\big), \\
    p_{\mathcal{M}_i}(y_t|y_{<t}, \mathbf{X},\mathbf{Q}) &= \frac{\exp(\mathbf{q}_{t-1})}{\sum_{j=1}^{|V|} \exp(\mathbf{q}_j)}\\
\end{split}
\label{next_word_prob}
\end{equation}
where $\mathbf{q}_{t-1}$ represents the output vector of the final linear layer of a language model. This formulation carries significant dependencies to visual tokens and language tokens. For the input image, if the visual tokens cannot represents the its important aspects, the language model cannot create connections with to input prompt. Since our goal is to compare and contrast different VLMs, we must demystify the visual and language understanding of the model and answer questions, such as: \textit{"Does the model correctly encode the image so that it preserve the semantic structure?"} or \textit{"Does the answer of the model faithfully reflect the content of the image?"} Therefore to answer these questions, we design the \textit{Analysis} step which is the core of our methodology.


\section{Visual–Linguistic Analysis of VLMs}
\label{sec:analysis}

Given a pool of candidate (base) VLMs as an ensemble, V3Fusion first performs the focal diversity-based ensemble pruning for two reasons: 
First, the diversity among base models improves the ensemble performance \cite{breiman1996heuristics, dietterich2000ensemble}. Second, as we add more models to the ensemble pool, it becomes more expensive to prompt each model, and the input length of the ensemble model increases. Thus, the model selection for an ensemble set is essential.

Consider a pool of $N$ base models, the total number of possible ensemble teams with size $S$ ($2\leq S \leq N$) is $2^{N}-N-1$. A key question is how to perform ensemble pruning efficiently. We argue that the smaller ensemble size and the higher ensemble diversity, the better the generation performance of the ensemble. Therefore we design two metrics, first is to measure visual understanding diversity among the models, and  the second, is to measure output diversity to capture linguistic relations. 

\vspace{4pt}

\subsection{Visual-Understanding-Diversity Metric}

To design the analysis that we described in Section \ref{sec:closer}, we must compare the visual understanding of each VLM. Therefore, during forward propagation of each VLM in the pool of $N$ models, we hook the visual-encoders $g_{\mathcal{M}_1}, \dots, g_{\mathcal{M}_N}$ and obtain the visual token embeddings. However, each model utilizes a different ViT with various structural differences, e.g., different patch sizes and pooling mechanisms, and transformations. Thereby, each model generates embeddings with different dimensions.  To address these challenges, we employ Central Kernel Alignment (CKA) \cite{kornblith2019similarity, cortes2012algorithms, raghu2021vision}, which enables quantitative comparison by looking at the relative relation between the embedding of samples extracted by different models. First, we take the mean-pool of vision token embeddings of model $i$, denoted by $\widetilde{\mathbf{h}}_i = \mathrm{MeanPool([\mathbf{h}_{I,1}, \dots, \mathbf{h}_{I,L'}])}$. We repeat for $K$ samples to obtain $\widetilde{\mathbf{X}}_i\in\mathbb{R}^{K\times d_i}$ and $d_i$ is the embedding dimension of model-$i$. Let $\widetilde{\mathbf{X}}_j$ be the embedding of model$-j$. Gram matrices of both embeddings are then computed as $\mathbf{K}=\widetilde{\mathbf{X}}_{i}\widetilde{\mathbf{X}}_{i}^{\top}$ and $\mathbf{L}=\widetilde{\mathbf{X}}_{j}\widetilde{\mathbf{X}}_{j}^{\top}$ capturing the internal similarity structure of each embedding space. Then, CKA compares the pairwise similarity by calculating:
\begin{equation}
    \mathrm{CKA}(\widetilde{\mathbf{X}}_i, \widetilde{\mathbf{X}}_j) = \frac{\mathrm{HSIC}(\mathbf{\mathbf{K}, \mathbf{L}})}{\sqrt{\mathrm{HSIC}(\mathbf{\mathbf{K}, \mathbf{K}})\mathrm{HSIC}(\mathbf{\mathbf{L}, \mathbf{L}})}},
\end{equation}
where $\mathrm{HSIC}$ is the Hilbert Schmidt Independence Criterion which is calculated as $\mathrm{HSIC}(\mathbf{K}, \mathbf{L}) = \frac{\mathrm{vec}(\mathbf{K}')\cdot \mathrm{vec}(\mathbf{L}')}{n-1}$. Here $\mathbf{K}'=\mathbf{H}\mathbf{K}\mathbf{H}$ and $\mathbf{L}'= \mathbf{H}\mathbf{L}\mathbf{H}$ are centered by $\mathbf{H}=\mathbb{I}_n-\frac{1}{n}\mathbf{1}\mathbf{1}^{\top}$ ensuring the invariance to isotropic scaling. CKA, also, invariant to orthogonal transformation of embeddings, enabling meaningful comparison and analysis of extracted sample embeddings.

\textbf{Focal-CKA:} Observe that, CKA is a pairwise similarity metric between two models, yet we have an ensemble set containing multiple models denoted by $\mathcal{E}_i = \{{\mathcal{M}_1, \dots, \mathcal{M}_S}\}$ where $S=|\mathcal{E}_i|$. Therefore we device \textit{Focal-CKA} that represents visual-understanding-diversity among the models score this candidate set. As shown in the second stage of Figure \ref{fig:ensemble}, we first obtain a CKA similarity matrix $\mathbf{S}\in[0,1]^{N\times N}$, where $s_{ij} = \mathrm{CKA}(\widetilde{\mathbf{X}}_i,\widetilde{\mathbf{X}}_j)$, $s_{ij} = s_{ji}$, and $s_{ii}=1$. Next, we select a focal model $\mathcal{M}_i$ inside the ensemble set and compute average pairwise CKA similarity, $\kappa^{focal}$ between focal model and the other models in the set. We repeat this process for each model in the ensemble set to obtain Focal-CKA:
\begin{equation}
\begin{split}
    \lambda^{focal}_{CKA}\left(\mathcal{E}_i\right)=1-\frac{1}{|\mathcal{E}_i|}\sum_{\mathcal{M}_i\in\mathcal{E}_i}{\kappa^{focal}(\mathcal{M}_i;\mathcal{E}_i)} \\
    \kappa^{focal}\left(\mathcal{M}_{i};\mathcal{E}_{i}\right)=\frac{1}{|\mathcal{E}_{i}|-1}\sum_{\mathcal{M}_j\in\mathcal{E}_{i} \setminus \mathcal{M}_i} s_{ij}\\
\end{split}
\label{eq:focal-cka}
\end{equation}
We subtract the average $\kappa$ from 1 to obtain a score representing visual diversity, where 1 represents the most dissimilar and 0 represents the most similar visual understanding. Most importantly, to represent the error diversity and its visual disagreement among models, we use the episodes in which the focal model made incorrect decisions (negative episodes) to calculate the Focal-CKA matrix.

\subsection{Focal Negative Correlation \& Focal Diversity}
The focal negative correlation metric, $\rho^{focal}$ is used to quantify the level of error diversity among the component models of an ensemble concerning each model within the ensemble. The focal diversity metric $\lambda^{focal}_{error}$ 
is used to quantify the general error diversity of the ensemble by taking into account all focal negative correlation scores of an ensemble. 
Similarly, let $\mathcal{E}^i$ denote a VLM ensemble composed of $S$ models. We choose one of the $S$ base models each time as the focal model to compute the focal negative correlation score of this ensemble, denoted as $\rho^{focal}(\mathcal{M}_i; \mathcal{E}^S)$. We define the focal diversity of this ensemble team by the average of the $S$ focal negative correlation scores.
The procedure of computing the focal negative correlation score of $\rho^{focal}$ is as follows: 
(i) select a base model among the set of $S$ base models as the \textit{focal} model, (ii) take all the validation episodes that the focal model has failed and calculate the focal negative correlation score, (iii) repeat the previous steps until all $S$ focal negative correlation scores are obtained. $\{\rho^{focal}_1, \dots, \rho^{focal}_S\}$, and (iv) compute the average over the scores to obtain the focal diversity of ensemble $\mathcal{E}^S$, denoted by $\lambda^{focal}(\mathcal{E}^S)$:
\begin{equation}
{\small
\begin{split}
\lambda^{focal}_{\mathrm{error}}(\mathcal{E}_i)=\frac{1}{S}\sum_{\mathcal{M}_i \in \mathcal{E}_{i}} \rho^{focal}(\mathcal{M}_i; \mathcal{E}_i)\\
\rho^{focal}(\mathcal{M}_i; \mathcal{E}_i ) = 1 - \frac{P(2)}{P(1)}\\
P(2)=\sum_{j=1}^{S}\frac{j(j-1)}{S(S-1)}p_j, \;
P(1)=\sum_{j=1}^{S}\frac{j}{N}p_j
\end{split}
\vspace{-8pt}
}
\end{equation}

Here, $p_i$ is the probability that $i$ models fail together on a randomly chosen episode. We calculate as $p_i={n_i}/{K}$ where $n_i$ is the total number of episodes that $i$ number of models failed together on the observation set $\mathcal{D}$ and $K$ is the total number of validation episodes. The term $P(2)$ represents the probability of two randomly chosen models simultaneously failing on an episode, while the denominator, $P(1)$, represents the probability of one randomly chosen model failing on an episode. The terms beneath $p_j$ values are the probability of the chosen model being one of the failures. For example, when $S=3$, there are three cases of model failures; one, two, or three models can fail simultaneously. If one model fails, the chance of selecting the failed model is $1/3$. Similarly, for two models, it is $2/3$, and for three models, it is $1$.
In the case of minimum diversity, the probability of two randomly chosen models failing together comes down to the probability of one of them failing, which makes the fraction term equal to 1 and $\rho^{focal} = 0$. Similarly, in the case of maximum diversity, there are no simultaneous failures. Hence, the nominator equals 0 and $\rho^{focal} = 1$. The definition of error changes according to the type of task and its solution set $y$. For the MCQs and OEQs, the errors are inequalities between the prediction of the model and the label; for the GQs, the errors are missed 1-grams between the prediction and the label. Thus, the focal diversity captures member models that are not correlated solely by their error diversity.

\vspace{-5pt}
\section{Ensemble Pruning with Focal Diversity}
\label{sec:ensemble_pruning}
\vspace{-5pt}

\begin{figure}[t]
    \centering
    \begin{minipage}{0.65\columnwidth}
    \centering
        \begin{subfigure}{0.48\textwidth}
        \centering
        \includegraphics[width=\linewidth]{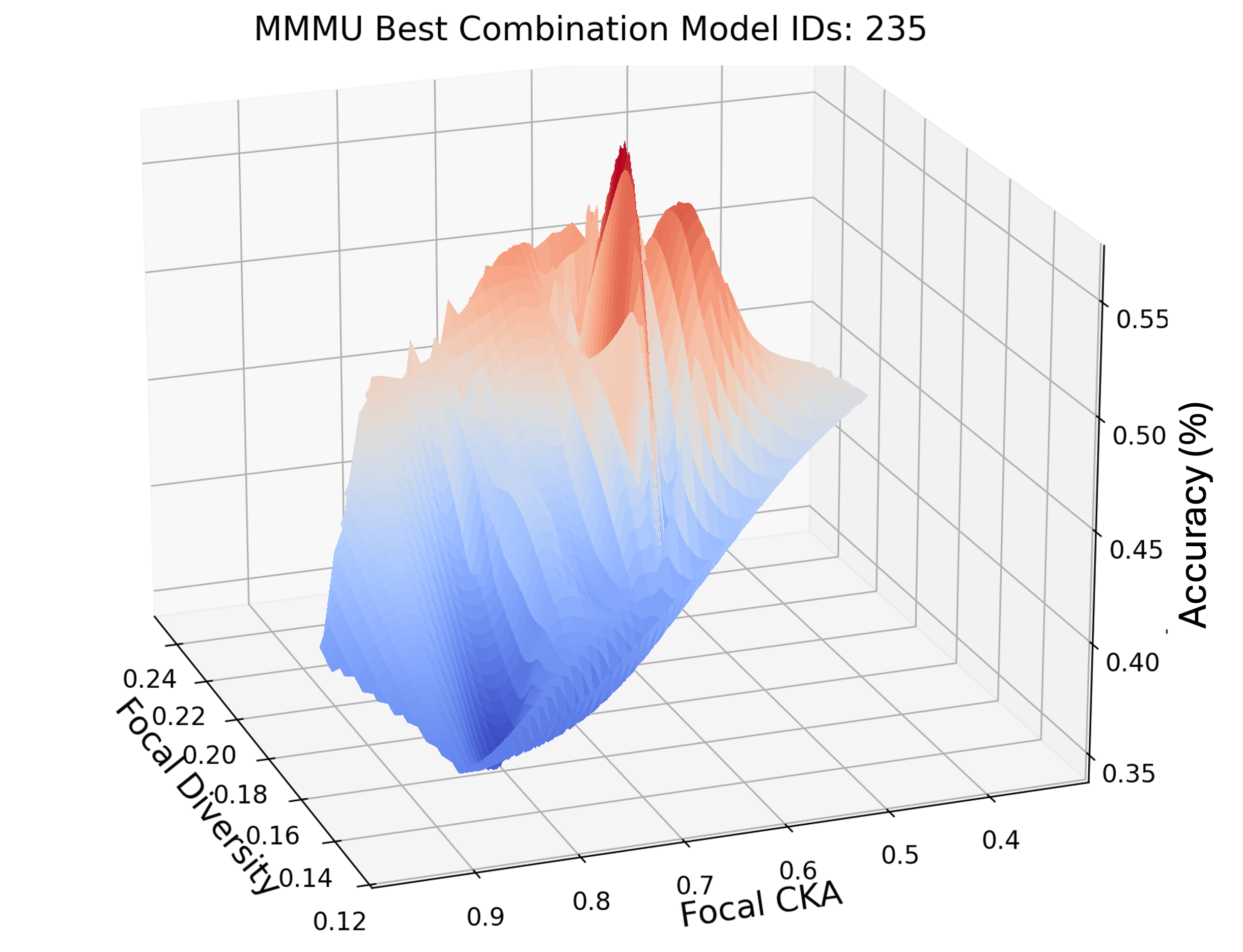}
        \caption{}
        \label{fig:focal_first}
        \end{subfigure}
        \hfill
        \begin{subfigure}{0.48\textwidth}
            \centering
            \includegraphics[width=\linewidth]{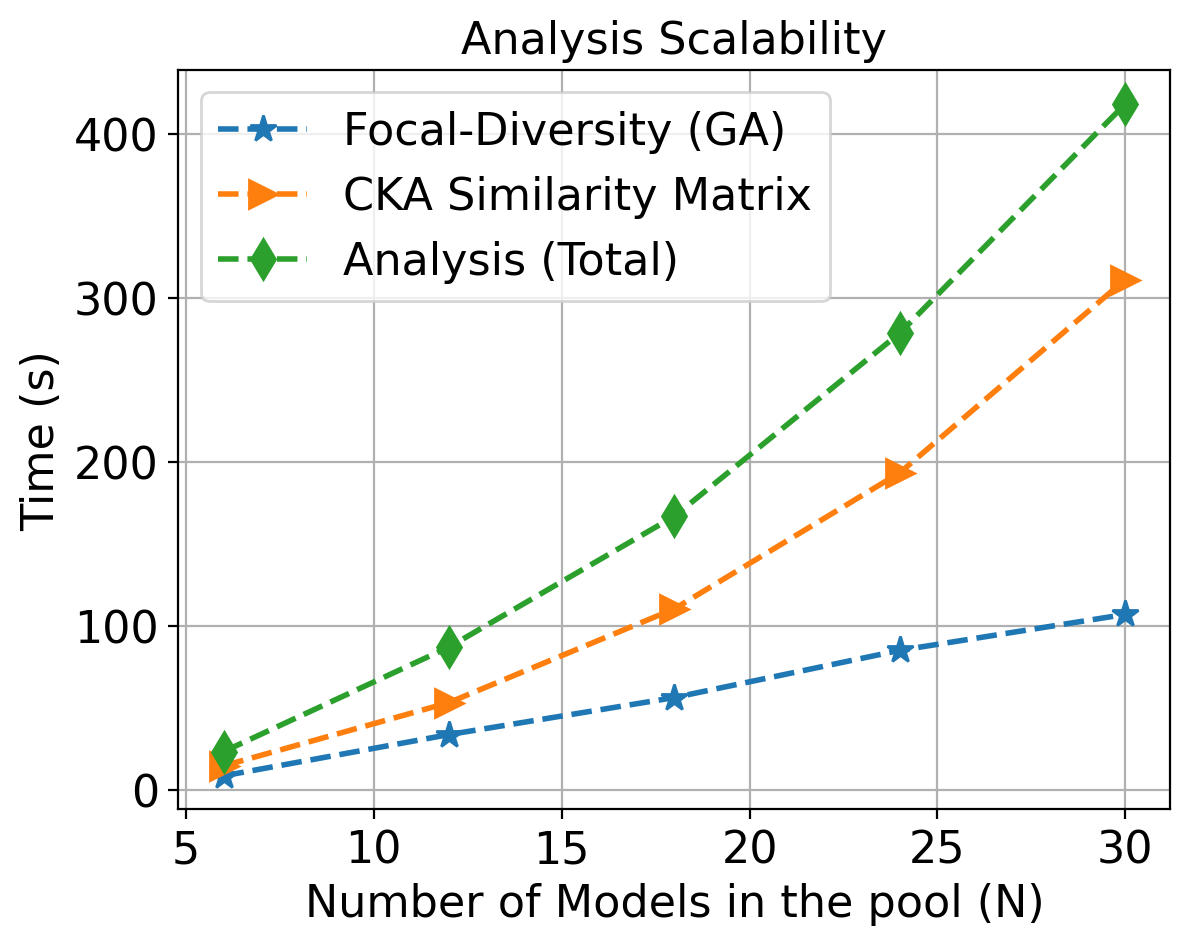}
            \caption{}
            \label{fig:scalibility}
        \end{subfigure}
        \vspace{-8pt}
        \caption{(a) All candidate ensemble teams from the model pool are plotted with their Focal Diversity, Focal-CKA, and Plurality Voting metrics for MMMU dataset. (b) Scalability of the analysis stage with breakdown.}
    \end{minipage}
    \hfill
    \begin{minipage}{0.33\columnwidth}
    \captionof{table}{Brute Force (BF) and Genetic Alg. (GA) pruning comparison with time reduction percentage.}
    \begin{adjustbox}{width=1\textwidth, center}
    \centering
    \small
    \begin{tabular}{l c c c}
      \hline
      \multirow{2}{1.5cm}{VLM Count} & \multicolumn{2}{c}{Time(s)} & \multirow{2}{1.5cm}{\centering Reduction (\%)$\uparrow$}\\
      \cline{2-3}
      & BF$\downarrow$ & GA$\downarrow$ & \\
      \hline
      $N=5$ & 9.4 & 9.9 & $-5.31$ \\
      $N=10$ & 228.2 & 24.5 & \textcolor{darkgreen}{89.26} \\
      $N=15$ & 508.99 & 41.8 & \textcolor{darkgreen}{91.78} \\
      $N=20$ & 16201.51 & 54.9 & \textcolor{darkgreen}{99.66} \\
      \hline
    \end{tabular}
    \end{adjustbox}
    \label{fig:speed}
    \end{minipage}
\end{figure}

Figure~\ref{fig:focal_first} shows the surface created by Focal-CKA, Focal-Diversity, and the accuracy of every ensemble combination for a given pool of $N=6$ base models with MMMU \cite{yue2024mmmu}. In Figure \ref{fig:correlation} at Appendix, we plot them separately. We make three observations: (i) the Focal-CKA and Focal-Diversity metric is correlated with the model performance, (ii) there are multiple sub-ensemble teams of size 2-4 that outperform the largest ensemble of size $6$, and (iii) a majority of the smaller ensemble teams also outperform the best-performing individual model in the base model pool.

The best combination we need to select is lying at the summit of the surface in Figure \ref{fig:focal_first}. However, we need to compute the focal diversity scores for all $2^{N}-N-1$ sub-ensemble teams when given a pool of $N$ base models. The brute force (BF) approach requires computing the focal diversity for each candidate ensemble of size $S$ ($2\leq S\leq N$). For $N=20$, we need to compute the focal diversity score for all $1,048,555$ candidate ensemble teams. To speed up this process, we leverage the Genetic Algorithm (GA) \cite{mirjalili2019genetic}, which significantly reduces the time required to reach the optimal combination (the summit). The process is shown at the third step in Figure \ref{fig:ensemble}. Table \ref{fig:speed} shows a comparison. We used the average of focal scores as the fitness function denoted by $r_{\mathrm{fit}}(\mathcal{E}_i)=(\lambda^{focal}_{CKA} + \lambda^{focal}_{\mathrm{error}})/2$. For a pool of $N=20$ base models, we complete the focal diversity-based ensemble pruning in under a minute, achieving a 5-order-of-magnitude speedup and $99.66\%$ reduction in latency. Overall, as shown in Figure \ref{fig:scalibility}, the analysis stage scales sup-linearly.

\section{Action: Combining VLMs Outputs}
\label{sec:fuse_summary}

As we described in Section \ref{sec:problem_definition}, there are two types of questions that we aim to ensemble, MCQs and OEQs. We adopt two different ensemble strategies for each one, as shown at the Stage-4 in Figure \ref{fig:ensemble}.

\textbf{For an MC question}, following \cite{eval-harness, open-llm-leaderboard}, we aggregate the probabilities of the tokens creating the whole choice to compute the probability of an answer. After repeating the procedure for all the choices, we obtain the probability distribution over the choices, denoted by $\mathbf{q}=[q_1, \dots, q_m]$, where $q$ represents the probability of a choice and $m$ is the number of choices. Following \cite{tekin2024llm}, we adopt Multi-layer Perceptron (MLP) containing multiple layers of fully connected weights with sigmoid activation functions for fusing outputs of VLMs, which we call \textit{V3Fusion-MLP}. At the final layer, the ensemble performs softmax to produce the output probability:

\begin{equation}
    \tilde{y} = \mathrm{softmax}(\mathbf{W}_{H} (\dots \sigma(\mathbf{W}_{1}[\mathbf{q}_{1}, \dots,\mathbf{q}_{N}])\dots)),
    \label{eq:fusion}
\end{equation}
where $H$ is the number of layers. The first layer takes the concatenation of the probabilities as the input, i.e., $\mathbf{W}_1\in\mathbb{R}^{(mN)\times d}$ where $d$ is the input dimension of the second layer. We train the model by decreasing the cross-entropy loss $\mathcal{L}_{\mathrm{vote}}(y, \tilde{y}) = -\sum_{i=1}^{m} y_i \log(\tilde{y}_i)$ using SGD and observation dataset $\mathcal{D}$.

\textbf{For an OE question}, we design the V3Fusion learn-to-ensemble by implementing Longformer Encoder-Decoder (LED) \cite{beltagy2020longformer} training method (V3Fusion-LED for short) with two motivations. First, V3Fusion-MLP is not applicable without relaxing the definition of equality between different solutions and it is not generative. Second, our goal with V3Fusion-LED is to create an ensemble learner that applies to all types of tasks and generates its own output \cite{tekin2024llm}. V3Fusion-LED is a sequence-to-sequence (seq2seq) model with encoder-decoder architecture~\cite{jiang2023llm} by concatenating the outputs of the component models of a chosen ensemble of $S$ base models with the input query and generating the final output of V3Fusion. It exploits sparse attention and global attention such that we can increase the context length up to 16396 tokens with 149 million parameters, and utilize a small training dataset. We give the details of the model in Appendix \ref{sec:led_details}.

\section{Verification \& Rectification}
\label{sec:verify}

There are two types of uncertainties one can model, \textit{aleatoric} (data) uncertainty and \textit{epistemic} (model) uncertainty \cite{kendall2017uncertainties}. In this section, we leverage the representativeness of multiple VLMs to calculate the system's epistemic uncertainty. Bayesian Neural Networks (BNNs) \cite{blundell2015weight, graves2011practical} and {\small DEEP ENSEMBLES} \cite{lakshminarayanan2017simple} have shown the standard approaches to uncertainty decomposition. Following \cite{malinin2018predictive}, the model uncertainty can be described by posterior distribution over the model parameters $p(\mathbf{\theta}|\mathcal{D})$. Then, Mutual Information (MI) $I[y, \mathbf{\theta}|\mathbf{x}]$ between the categorical label $y$ and parameters of the model $\mathbf{\theta}$ measures the dependency between the predicted label and the model parameters, reflecting the variability among ensemble members $\{P(y|\mathbf{x}, \theta^{(i)})\}_{i=1}^{N}$ which assess uncertainty in predictions due to model uncertainty. Formally one can calculate the MI by the difference between total uncertainty data uncertainty:
\begin{equation}
    \underbrace{I[y, \mathbf{\theta}|\mathbf{x}]}_\text{Epistemic} = \underbrace{\mathcal{H}[\mathbb{E}_{p(\mathbf{\theta}|\mathcal{D})}[P(y|\mathbf{x},\theta)]]}_\text{Total Uncertainty} - \underbrace{\mathbb{E}_{p(\mathbf{\theta}|\mathcal{D})}[\mathcal{H}[P(y|\mathbf{x},\theta)]]}_\text{Aleatoric Uncertainty}
\end{equation}
where $\mathcal{H}$ represents the entropy among the logits of the models and here we model the total uncertainty by calculating the entropy of the ensemble model $\mathcal{H}_{ens}[P(y|\mathbf{x},\theta)]$ and aleatoric uncertainty by taking the average of base model entropy scores. The epistemic uncertainty indicates that when the model is unsure, it hasn't learn the data distribution well, and eventually will help us to prevent hallucination. Unlike the literature, we model epistemic uncertainty using heterogeneous models with diverse architectures and vocabularies. This variety reflects the true disagreement and better approximates a posterior over models to capture the knowledge gaps among base models more effectively. 

\begin{algorithm}[t]
\caption{Adaptive Epistemic Uncertainty}
\label{alg:adaptive}
\small
\begin{algorithmic}[1]
\Require Uncertainties $\{e_i\}_{i=1}^{N}$, significance threshold $\alpha$
\State $\mu, \sigma \leftarrow \mathrm{FitGaussian}(\{e_i\})$ \Comment{Fit single Gaussian}
\State $\mathrm{Log}L_1 \leftarrow \sum_{i=1}^N \log \mathcal{N}(e_i \mid \mu,\sigma^2)$
\State $(\pi, \mu_1,\sigma_1,\mu_2,\sigma_2) \leftarrow \mathrm{FitGMM}(\{e_i\})$ \Comment{Fit 2-comp. GMM}
\State $\mathrm{Log}L_2 \leftarrow \sum_{i=1}^N \log[\pi\,\mathcal{N}(e_i\mid\mu_1,\sigma_1^2) + (1-\pi)\,\mathcal{N}(e_i\mid\mu_2,\sigma_2^2)]$
\If{$\mathrm{Log}L_2 - \mathrm{Log}L_1 > \alpha$}
    \State Predict cluster labels $c_i$
    \State $G_0 \gets \{e_i : c_i=0\}$,\quad $G_1 \gets \{e_i : c_i=1\}$
    \State $\tau \gets \min\{\max(G_0),\max(G_1)\}$
\Else
    \State $t \leftarrow \mu + 2\sigma$
\EndIf
\State \Return Threshold $t$
\end{algorithmic}
\end{algorithm}

An unanswered question is how to select the uncertainty threshold $\tau$ dynamically. The incoming data to the system can comprise both uncertain and certain samples, resulting in two distributions that we observe are centered at different positions. Therefore, we apply the likelihood ratio test to two probability density models and select the one that best represents the observed epistemic uncertainty data \cite{bishop2006pattern, burnham2004multimodel, anonymous2025gradshield}. The algorithm \ref{alg:adaptive} illustrates adaptive threshold selection, where we either select a two-component Gaussian Mixture Model or a single Gaussian, depending on the significance threshold $\alpha$, which can be 10 corresponding to $0.01$ p-value.

We reject the samples that have higher Epistemic Uncertainty than the threshold value $\tau$ and and accept the ones that have lower. For the rejected samples, we predict by averaging all the base model logits and return the maximum choice as the final output. This way our methodology can decide on the final answer without the need of external system or supervision making it self-sustained closed system. 

\section{Experiments}

\label{sec:experiments}

We validate the effectiveness of V3Fusion through extensive evaluations on benchmarks with two question types, multiple-choice and open-ended. We demonstrate that V3Fusion outperforms both the best individual V3Fusion models and fine-tuned VLMs for the ensemble task. Surprisingly, our V3Fusion works in the cases where the majority of the candidate models make errors. Lastly, we provide the ablation study to show the gain by focal error-diversity optimized ensemble pruning. The experiments are performed on four popular VQA benchmark datasets: MMMU\cite{yue2024mmmu}, MMMU-Pro \cite{yue2024mmmu_pro}, A-OKVQA \cite{schwenk2022okvqa}, and OCR-VQA~\cite{mishra2019ocr}. We give all the experimental setup details in the Appendix-A.

\begin{table*}[hbt!]
\vspace{-12pt}
\caption{The mean accuracy of base models and V3Fusion in OKVQA, MMMU, and MMMU-pro datasets. We create the ensemble sets using focal-diversity on $^\dagger$ OKVQA, $^{*}$  MMMU, $^\ddagger$ MMMU-Pro, and $^{**}$ OCR-VQA.}
\vspace{-5pt}
  \begin{adjustbox}{width=1\textwidth, center}
    \centering
    \small
    \begin{tabular}{p{3.5cm} c c c c c c c}
        \hline
        \multirow{2}{*}{Model Name} & \multirow{2}{*}{Model ID} & A-OKVQA$^\dagger$ & MMMU (Val)$^{*}$ & MMMU-Pro$^\ddagger$ & \multicolumn{3}{c}{OCR-VQA$^{**}$}\\
        \cline{3-8}
        & & Acc (\%)$\uparrow$ & Acc (\%)$\uparrow$ & Acc (\%)$\uparrow$ & BLEU-1$\uparrow$ & EM (\%)$\uparrow$ & F1(\%)$\uparrow$ \\
        \hline
        LlaVA-v1.6-Vicuna-13b & 1 & 83.4 & 37.02 & 33.69 & 78.07 & 67.71 & 79.60 \\
        LlaVA-v1.6-Vicuna-7b & 2 & 82.62 & 35.9 & 32.81 & 78.93 & 69.17 & 80.44 \\
        DeepSeek-VL2-Tiny & 3 & 76.68 & 39.01 & 34.61 & 38.30 & 29.32 & 41.71 \\
        DeepSeek-VL2-Small & 4 & 80.08 & 39.75 & 38.00 & 54.60 & 41.71 & 57.25 \\
        Qwen2.5-VL-7b-Instruct & 5 & 87.24 & 51.55 & 46.98 & 83.34 & 72.00 & 84.80 \\
        Intern-VL2-8b & 6 & 85.32 & 51.30 & 43.09 & 52.36 & 45.55 & 57.09 \\
        \hline
        V3Fusion-MLP & $123^{\dagger} \mid 235^{*} \mid 234^{\ddagger}$ &  88.31 & 55.07 & 47.34 & - & - & - \\
        V3Fusion-LED & $123^{\dagger} \mid 235^{*} \mid 234^{\ddagger} \mid 456^{**} $ &   87.86 & 51.30 & 46.26 & \textbf{86.24} & 71.91 & \textbf{86.82} \\
        V3Fusion-Rectify & $123^{\dagger} \mid 235^{*} \mid 234^{\ddagger} \mid 456^{**}$ & \textbf{89.17} & \textbf{56.09} & \textbf{49.27} & 85.71 & \textbf{72.08} & 86.57 \\
        \hline
        Relative Gain (\%) & - & \textcolor{darkgreen}{$+2.12$} & \textcolor{darkgreen}{$+8.09$} & \textcolor{darkgreen}{$+4.87$} & \textcolor{darkgreen}{$+3.48$} & \textcolor{darkgreen}{$+0.11$} & \textcolor{darkgreen}{$+2.85$}  \\
        \hline
    \end{tabular}
    \end{adjustbox}
    \label{table:gsm8k}
\end{table*}

\begin{figure*}[t]
    \centering
    \includegraphics[width=1\textwidth]{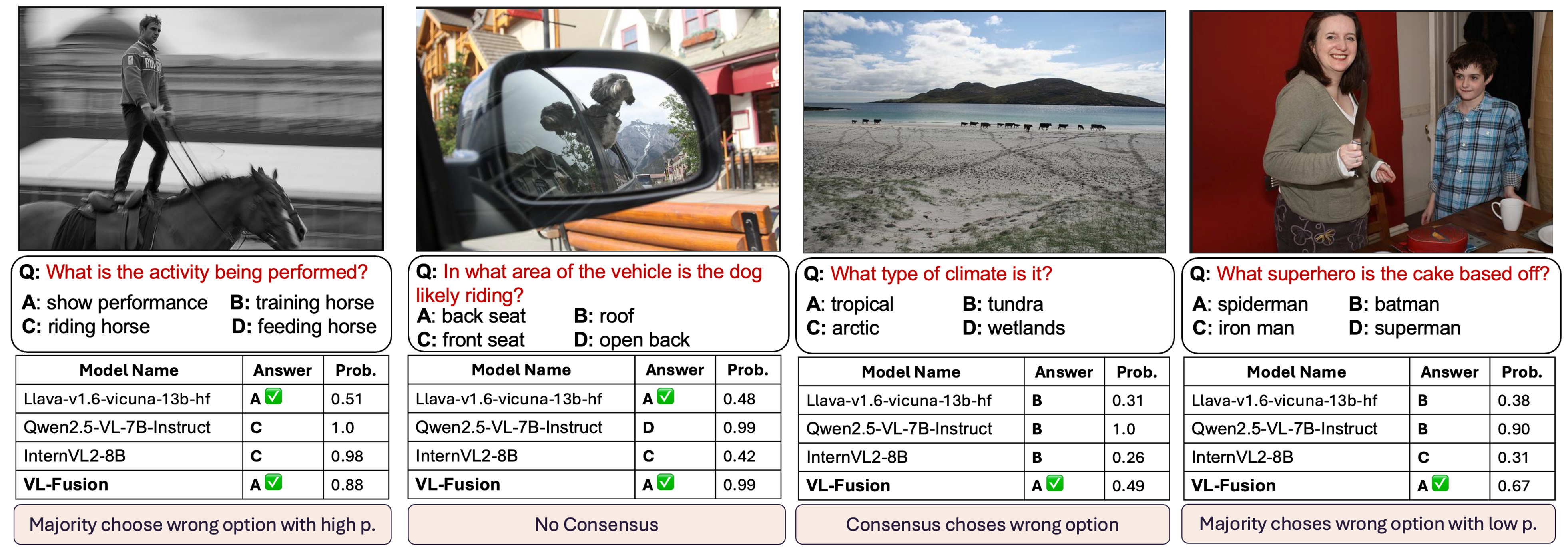}
    \caption{Four examples are used to illustrate the superior performance of V3Fusion compared to existing popular methods. It demonstrates cases where V3Fusion can achieve the correct output, even when the base models fail to reach a consensus or agree on an incorrect option.}
    \label{fig:graphs}
    \vspace{-10pt}
\end{figure*}

\textbf{Performance of V3Fusion:} 
Table \ref{table:gsm8k} presents experiments on datasets, comparing the scores of each base model in the pool with those of the ensemble learners V3Fusion-MLP and V3Fusion-LED, plus their rectified version with V3Fusion-Rectify. The Model IDs of V3Fusion denote the models in the ensemble set, which is selected by focal diversity pruning. In the MCQ tasks (MMMU, MMMU-Pro, and A-OKVQA), V3Fusion-MLP reaches the best performance by surpassing the best-performing model Qwen2.5-7B by 5\%, 3\%, and 2\%. Whereas V3Fusion-LED shows the same performance as the best-performing model. The reason is that when V3Fusion-MLP is making a decision, it considers the probabilities assigned to the candidate outputs and observes the pattern for each candidate model. As shown in Figure \ref{fig:graphs}, V3Fusion-MLP can reach the correct answer when the majority or even all the models are giving an incorrect decision. For example, in the third sample in \ref{fig:graphs}, models show consensus at choice B, but the V3Fusion selected the correct decision. When we analyze the probabilities of choices assigned by the models, we observe that Llava-1.6v assigned 0.29 probability to option A. This indicates that, despite Llava-1.6 being the third-best performing model, it contributes valuable information that the V3Fusion model leverages to achieve the correct final decision. On the other hand, the V3Fusion-LED model performs learnable consensus through summarization, which requires reasoning and thinking to generate high-quality output. Hence, it is more suitable for open-ended VQA tasks such as OCR-VQA, where it improves the best model's F1 score by $2.2\%$. V3Fusion-LED successfully detects the information gathered by each base model and generates the correct output. Each model has its own expertise due to its training dataset's coverage and its learning capabilities, which the visual and language encoding models facilitate. V3Fusion-LED can combine answers from multiple VLMs via learn to combine by summarization by exploiting the complementary wisdom of individual base models of an ensemble.

\begin{figure*}[t]
    \centering
    \includegraphics[width=\linewidth]{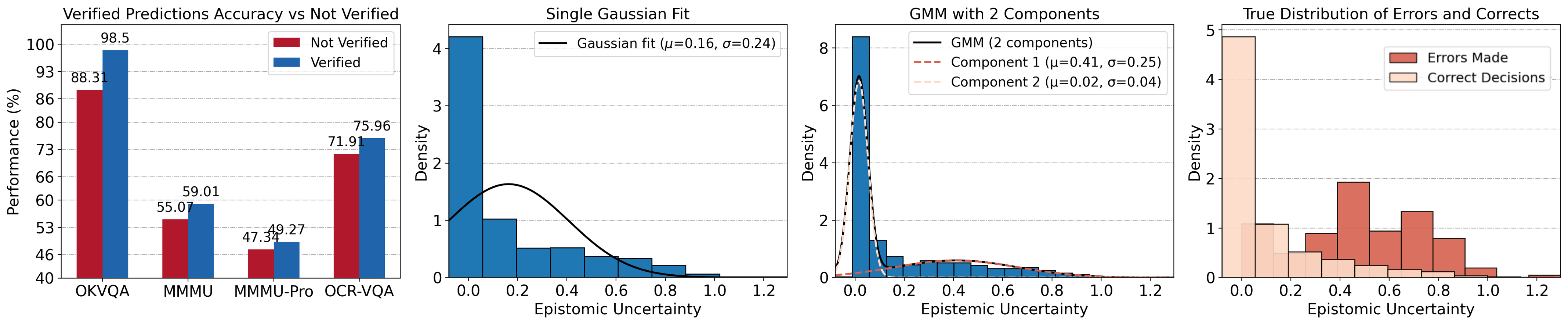}
    \caption{We illustrate the threshold selection process and the resulting boost by the Vision Verification. The first figure on the left shows the average accuracy of the verified predictions for each dataset compared to the non-verified predictions. The second and third plots show the single- and two-component Gaussian fits to the Epistemic Uncertainty, the Algorithm \ref{alg:adaptive} selects threshold as $\tau=0.1315$. }
    \label{fig:threshold}
    \vspace{-10pt}
\end{figure*}

\textbf{Boost by the Rectification Stage:} V3Fusion-Rectify boosts the performance even further in each dataset and obtains the highest score. The first plot on the left of Figure \ref{fig:threshold} shows the accuracy comparison between the verified samples by the Adaptive Epistemic Uncertainty algorithm. Almost all the samples that are verified in the OK-VQA dataset are correct, and we observe 4-5\% boost in other datasets. We observe minor improvements in the results of the MLP and LED models with rectification, indicating that the rejected samples are significantly more challenging and require larger models to handle. The following three plots in Figure \ref{fig:threshold} show the threshold selection for the OK-VQA dataset. It shows that a single Gaussian cannot represent the uncertainty distribution, so a 2-component GMM was selected and the threshold value $\tau=0.1315$ was found. The last plot shows the true uncertainty distribution on correct and incorrect samples. We observe that the selected threshold value represents a great cut-off value.

\textbf{Performance Against Prior Work:} We further evaluate V3Fusion by comparing it with three different lines of work in Table \ref{tab:ensemble_methods}. LLM-Blender's Pair Rank \cite{jiang2023llm} is a supervised ensemble approach. RouteLLM \cite{ong2024routellm} is a routing algorithm that selects the best-suited model for the query. Multi-Agent Debate (MAD) \cite{du2023improving} is a collaboration method where agents debate for multiple turns until a consensus is established or the judge decides. In all the datasets, V3Fusion reaches the best performance with the second-lowest inference time. While the other methods consider only the language output to decide on the best model selection, V3Fusion utilizes the visual modality in model selection and learns to combine the wisdom of each model.


\begin{table}[t]
\centering
\captionof{table}{VL3Fusion against routing, collaboration, and ensemble baselines.}
\vspace{-8pt}
\begin{adjustbox}{width=0.73\textwidth, center}
\begin{tabular}{l l c c c c c}
    \hline
    \multirow{2}{*}{Method} & \multirow{2}{1.3cm}{Infer. Time (s)} & \multicolumn{3}{c}{OCR-VQA} & MMMU & A-OKVQA \\
    \cline{3-6}
    & & BLUE-1 & EM & F1 & Acc (\%) & Acc (\%)\\
    \hline
    LLM-Blender (Pair R.)\cite{jiang2023llm} & 19.1s & 49.93 & 30.58 & 43.95 & 22.36 & 70.56 \\
    RouteLLM (SWRank)\cite{ong2024routellm} & 0.0158s & 81.69 & 70.94 & 83.27 & 30.68 & 81.83 \\
    RouteLLM (Random)\cite{ong2024routellm} & 0.0007s & 80.73 & 69.81 & 82.23 & 26.45 & 68.90 \\
    MAD w/ Consensus\cite{sivakumaran2025dart}& 21.84s & - & - & - & 53.2 & 57.4 \\
    MAD w/ Judge\cite{sivakumaran2025dart}& 27.92s & - & - & - & 54.8 & 59.7 \\
    \hline
    V3Fusion &  3.28s & \textbf{86.24} & \textbf{71.91} & \textbf{86.82} & \textbf{56.09} & \textbf{89.17} \\
    \hline
\end{tabular}
\end{adjustbox}
\label{tab:ensemble_methods}
\vspace{-8pt}
\end{table}




\begin{table}[t]
    \captionof{table}{We compare our approach with the supervised fine-tuned (SFT) VLMs with instruction-style ensemble prompt.}
    \vspace{-6pt}
    \begin{adjustbox}{width=0.6\textwidth, center}
    \centering
    \begin{tabular}{l c c c c}
        \hline
        \multirow{2}{*}{Method}  & \multirow{2}{*}{ID} & \multicolumn{3}{c}{OCR-VQA} \\
        \cline{3-5}
        & & BLEU-1$\uparrow$ & EM (\%)$\uparrow$ & F1$\uparrow$ \\
        \hline
        LlaVA-v1.5-Vicuna-7b (SFT) & 123456 & $69.25$ & $58.98$ & $71.24$ \\
        Qwen2-7B-VL-Instruct (SFT) & 123456 & $84.99$ & $75.27$ & $86.49$ \\
        Qwen2.5-7B-VL-Instruct (SFT) & 123456 & $85.65$ & $75.57$ & $87.04$\\
        \hline
        V3Fusion-LED & 456 & $\mathbf{87.79}$ & $74.91$ & $\mathbf{88.83}$ \\
        \hline
    \end{tabular}
    \end{adjustbox}
    \label{table:comparison_ens}
    \vspace{-15pt}
\end{table}

\textbf{Finetuning Base Models:}
We also compare the performance of finetuned base models and V3Fusion-LED in the generative tasks in Table \ref{table:comparison_ens}, using the same training data. We select Qwen2.5-7B, Qwen2-7B, and Llava-v1.5 as visual-language models to ensemble the candidate outputs generated by the base models. We used the same prompt, incorporating candidate model outputs for the Visual-Question pair, and fine-tuned the model using supervised fine-tuning (SFT) with LoRA. Training was stopped after three consecutive steps without a validation performance improvement. As shown in Table \ref{table:comparison_ens}, even though V3Fusion-LED uses a smaller number of parameters, it outperforms fine-tuned versions of the base models. Particularly, V3Fusion-LED has a shorter training time ($35$ mins) compared to VLM training ($1.47$ hours for Qwen2.5-7B). This occurs because these models use an image encoder and text decoder architecture designed to learn image-text relationships. As a result, fine-tuning them for summarization is less effective than using sequence-to-sequence models. We additionally fine-tune the base models individually (without the ensemble objective) and compare their performance and training time on OCR-VQA and MMMU against V3Fusion (Table \ref{tab:sft}, Appendix). V3Fusion showed significantly high performance in MMMU with half the training time.

\textbf{Ablation Studies:} We observe the gains from stages 3, 4, and 5 by conducting an ablation study in Table~\ref{tab:ablation}. The gains from pruning-only (with plurality voting as the final decision maker) and fusion-only (without pruning, i.e., all 6 models combined with the MLP) are inconsistent. On the other hand, V3Fusion with Pruning, Fusion, and Rectification stages provides the highest gain in all three datasets. This shows that the system reaches its highest performance when all three stages are active. Moreover, Table \ref{tab:metrics} shows scores of V3Fusion when we use popular pairwise error correlation metrics at the pruning stage. Compared to Focal-Diversity, these metrics underperform in both datasets. Therefore, Focal-Diversity is more capable of distinguishing the best model combination for given tasks. We perform efficiency analysis regarding the latency and data sensitivity of V3Fusion in Table \ref{tab:base_models} and \ref{tab:train_perc} in the Appendix-D.

\begin{figure}[t]
    \centering
    \begin{minipage}{0.49\columnwidth}
    \captionof{table}{Gains obtained by each phase individually for A-OKVQA, MMMU, and MMMU-Pro datasets.}
    \begin{adjustbox}{width=\textwidth, center}
        \begin{tabular}{l c c c}
            \hline
            Phases & A-OKVQA & MMMU & MMMU-Pro \\
            \hline
            Pruning Only & $+1.07$ & $+1.36$ & $-1.79$ \\
            Fusion Only & $-1.48$ & $+3.4$ & $+0.55$ \\
            V3-Fusion & $+\mathbf{1.93}$ & +$\mathbf{4.54}$ & +$\mathbf{2.29}$ \\
            \hline
        \end{tabular}
        \end{adjustbox}
        \label{tab:ablation}
    \end{minipage}
    \hfill
    \begin{minipage}{0.49\columnwidth}
    \captionof{table}{Focal diversity score against five popular pairwise error correlation scores over MMMU and OKVQA.}
    \begin{adjustbox}{width=\textwidth, center}
    \centering
    \begin{tabular}{p{1.3cm} p{0.8cm} p{0.8cm} p{0.8cm} p{0.8cm} p{0.9cm} p{0.9cm} p{0.8cm}}
        \hline
        Dataset & Fleiss  & Corr. Coef \centering & Bin. Disag.\centering & Kappa\centering & Bin. Entor.\centering & Focal-Div. \\
        \hline
        MMMU  & 37.14 & 37.14 & 42.61 & 37.14 & 50.31 & \textbf{54.50} \\
        OKVQA  & 82.79 & 82.79 & 79.04 & 82.79 & 76.77 & \textbf{88.31} \\
        \hline
    \end{tabular}
    \end{adjustbox}
    \label{tab:metrics}
    \end{minipage}
\end{figure}

\vspace{-4pt}
\section{Conclusion}
\label{sec:conclusion}
\vspace{-4pt}
We have introduced V3Fusion, a novel framework for diversity fusion of vision-language models (VLMs). This paper makes three original contributions. First, our V3Fusion framework develops the error diversity metrics to capture the complimentary wisdoms of individual component VLMs with respect to erroneous inference results. By leveraging our focal error diversity measures, we construct a candidate ensemble surface for a pool of candidate VLMs, and apply a Genetic Algorithm to effectively prune out those candidate VLMs that do not add value to the performance improvement of our V3Fusion framework. Second, we develop a holistic learning approach to combining and resolving output inconsistencies among the VLMs of a VLM fusion team by training a V3Fusion model. Finally, we calculated the epistemic uncertainty of the system to reject or accept the generated samples with a dynamic adaptive threshold. We conduct extensive experiments to validate V3Fusion on four popular vision-language benchmarks and observe that V3Fusion outperforms the best-performing VLMs.

\section*{Acknowledgment}
This research is partially sponsored by NSF CISE grants 2302720 and 2312758, an IBM faculty award, a grant from the CISCO Edge AI program, a GTRI PhD Fellowship, and the research cyberinfrastructure resources and services provided by the Partnership for an Advanced Computing Environment (PACE) at the Georgia Institute of Technology. 

\bibliographystyle{splncs04}
\bibliography{main}

\clearpage
\setcounter{page}{1}
\appendix

\section{Experimental Setup}

{\bf Datasets.\/} The experiments are performed on four popular VQA benchmark datasets: MMMU\cite{yue2024mmmu}, MMMU-Pro \cite{yue2024mmmu_pro}, A-OKVQA \cite{schwenk2022okvqa}, and OCR-VQA~\cite{mishra2019ocr}. The first three are VQA datasets 
in multiple-choice question format. MMMU\cite{yue2024mmmu} dataset is designed to evaluate VLMs on massive multi-discipline tasks containing 30 different subjects. The questions are at the college level and contain multiple reasoning steps. The dataset only provides 5 development and 900 validation samples with the labels. However, the labels of the test split are hidden. MMMU-Pro dataset~ \cite{yue2024mmmu_pro}, on the other hand, is the robust version of MMMU and provides only a test dataset with labels. To perform training on fusion models, we used MMMU-Pro to train the model and evaluate it on MMMU validation data. For the MMMU-Pro, we used MMMU validation data and tested it on MMMU-Pro test data. There are 577 common samples between the MMMU-Pro test data and the MMMU-val data. We remove the intersections from the train data. A-OKVQA dataset~\cite{schwenk2022okvqa} provides visuals with questions that require reasoning to solve. The expected answer is either the choice from the multiple choices given or a direct answer. There are 17,056 training pairs, 1,146 validation pairs, and 6,70 test samples.  For the A-OKVQA dataset, we use 3,000 samples from its training set to train our V3Fusion models and use the validation set for evaluation. Similar to the MMMU dataset, the A-OKVQA benchmark did not share the test labels. Thus, we used a validation split as the test data.

The OCR-VQA dataset~\cite{mishra2019ocr} consists of book cover images, each paired with four open-ended questions for OCR tasks. For example, the question \textit{What is the title of the book?} requires the model to recognize the text in the image and provide the answer in text form. Therefore, the answer is open-ended. There are 166,202 training pairs, 20,731 validation samples, and 20,796 test samples. V3Fusion-LED can perform reasoning effectively for such open-ended VQA tasks. 

\textbf{Evaluation Metrics.} To evaluate the performance of VLMs and V3Fusion methods, we use accuracy on MMMU, MMMU-Pro and A-OKVQA datasets. For the open-ended OCR-VQA dataset, we use BLUE-1, Exact Match (EM), and F1 scores to evaluate and compare the performance of individual VLMs and V3Fusion methods.

\textbf{VLM selection for the candidate model pool.} In our model pool selection, we choose pre-trained individual VLMs based on the following three elements and their effect on performance: (i) size of the model (less than 16b), (ii) model variety, and (iii) being open-source. The size was the limiting factor for us since the modern LLMs are huge, and our experiments were conducted on a single H100 GPU of 80GB.

\textbf{V3Fusion Training Models/Algorithms.} 
V3Fusion-MLP model contains two fully connected hidden layers with 100 neurons and ReLU activations between the layers. The model weights start from Xavier initialization and converge in 500 epochs optimized by Adam. To implement V3Fusion-LED, we employ the Longformer-Encoder-Decoder \cite{beltagy2020longformer} model, which is initialized from BART weights \cite{lewis2019bart}. The model contains additional positional encoding matrices on top of BART weights, with a total of 161 million parameters. Since not every question contains the same number of choices, we pad the generated probability vectors from each model to match the maximum number of choices.

\textbf{Ensemble Pruning.} We selected $w_1=0.5$, $w_2=0.25$, and $w_3=0.25$ while scoring a candidate ensemble set with the fitness function to give more importance to the focal diversity and equal importance to Fleiss Kappa and Plurality voting accuracy in multiple-choice type datasets. In the OCR-VQA, only focal diversity is used for pruning. The genetic algorithm stops when the fitness function does not change for 100 consecutive generations.

\begin{figure}[hbt!]
    \centering
    \includegraphics[width=0.5\textwidth]{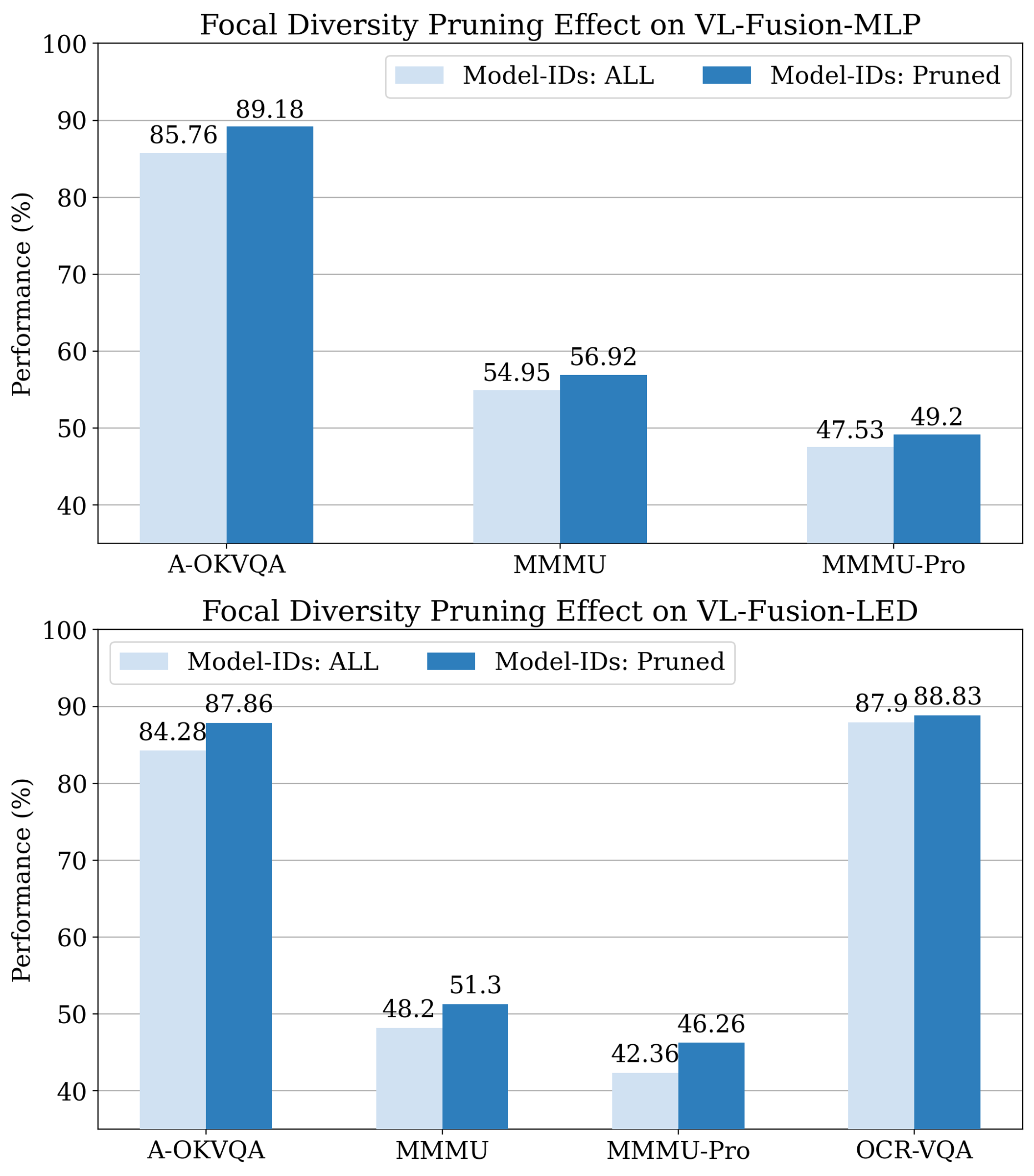}
    \vspace{-4pt}
    \caption{The effect of Focal-diversity Pruning on V3Fusion-MLP and V3Fusion-Weighted is shown for all the datasets.}
    \label{fig:focal_first}
\end{figure}

\section{Ablation Study}
To further observe the effect of the pruning and attention mechanisms, we execute two ablation studies in Figure \ref{fig:focal_first}. First, we ensemble all the models in the pool and compare their performances with the ensemble model selected by the pruning mechanism. As shown in both plots, the pruning improves the performance of V3Fusion-MLP and -LED in every task. Although there is less improvement in the V3Fusion-LED model in OCR-VQA, the pruned ensemble set is reaching a slightly better performance with fewer models. Given that the models are at a billion-scale parameter size, our approach significantly reduces computational redundancy compared to selecting all models.

\section{More Details for Fusing MCQs and OEQs}
\label{sec:led_details}

\textbf{For an MC question}, as proposed in \cite{hendrycks2020measuring}, the probabilities assigned to choices are obtained by calculating the probability of the choice's token using equation $\ref{next_word_prob}$. For instance, $p(w_t=\mathrm{A}|x, w_{<t})$ is calculated for choice A. However, a more popular methodology proposed by \cite{eval-harness} is used by the HuggingFace Leader Board \cite{open-llm-leaderboard} and also in our paper. We aggregate the probabilities of the tokens creating the whole choice to compute the probability of an answer. After repeating the procedure for all the choices, we obtain the probability distribution over the choices, denoted by $\mathbf{q}=[q_1, \dots, q_m]$, where $q$ represents the probability of a choice and $m$ is the number of choices.

As objective at Fusion Model, we want to find the best parameters $\theta = (\mathbf{W}_{1},\dots,\mathbf{W}_{H})$ to maximize the likelihood, which can be reduced to minimizing the cross-entropy loss on a dataset, which is the collection of probabilities for each component model. Thus, we split the dataset into training, validation, and test sets, use the training set to train the ensemble model, the validation set to stop training, and finally use the test set to calculate performance. In each iteration of training, the parameters are updated by minimizing the cross-entropy loss function:
\begin{equation}
\begin{split}
    \theta_{\mathrm{best}} &= \argmin_{\theta} \sum_{x, y \in \mathcal{D}^{\mathrm{train}}} \mathcal{L}_{\mathrm{vote}}(y, \tilde{y}), \\
     \tilde{y} &= f(\mathcal{M}_{1}(x), \dots, \mathcal{M}_N(x); \theta), \\
    \mathcal{L}_{\mathrm{vote}}(y, \tilde{y}) &= -\sum_{i=1}^{m} y_i \log(\tilde{y}_i).
\end{split}
\end{equation}
We use SGD to update the parameters at every iteration. The ensemble learner analyzes the probabilities assigned by each model and their confidence level. Thus, we train the learn-to-ensemble model to recognize patterns across the predictions of each component model efficiently. This allows the ensemble learner to learn to make the correct choice even in the absence of consensus, rather than blindly relying on consensus voting algorithms such as majority or plurality voting.

\textbf{For a OE question}, recall, $y^{(2)}=\{\omega_1, \dots, \omega_L\}$, where $L$ is the sequence length of the desired output. Each model in the pool generates the predicted sequence denoted by $\mathcal{M}_{i}(\mathbf{x}, \mathbf{I})=\{\hat{\omega}_1, \dots, \hat{w}_{T_i}\}=z_i$ and $L_i$ is the sequence length of the $i^{\mathrm{th}}$ model output which can be different than $L$. Let $\mathcal{Z} = \{z_1, \dots, z_N\}$ be the collection of candidates, VL3Fusion-LED models the conditional probability of $p(y|\mathbf{x}, \mathcal{Z})$. We give the input sequence, $\mathbf{x}_s$, to the seq2seq model in the format of $\mathbf{x}_s = \mathrm{concat}(x, z_1, \dots, z_N)$ and use special tokens as separators to indicate the beginning and end of the question or an answer. Consider an ensemble from 3 base models, the input below is sent to the V3Fusion-LED model: 
\begin{equation}
\small
 \begin{split}
 x_s = \mathrm{<boq>} x \mathrm{<eoq>} \mathrm{<boc1>} z_1 \mathrm{<eoc1>} \\ 
 \mathrm{<boc2>} z_2 \mathrm{<eoc2>}\mathrm{<boc3>}z_3\mathrm{<eoc3>}. 
 \end{split}
\end{equation}
We use distinct tokens to indicate which model each candidate belongs to. To reduce complexity and increase context length, we utilize the \textit{sliding window attention} pattern proposed by \cite{beltagy2020longformer}. To stress the relation between the question and each candidate's answer, we employ \textit{selective global attention} on the tokens of $x$ of the input question. The global attention is the standard self-attention by scoring each token against every other token. With the sliding and global attention mechanism, we increase the context window length, reduce the computational complexity, and improve the performance.

Overall, V3Fusion-LED is optimized by finding the best model parameter $\phi$ that will maximize the joint distribution over the target tokens $p(y|x, z_1, \dots, z_N;\phi)$. It performs auto-regressive generation using the following cross-entropy loss $
\mathcal{L}_{sum} = -\sum_{t=1}^{T}\log p(w_t|w_{<t-1}, x, \mathcal{Z};\phi)$
We use SGD to perform updates on the parameters in each iteration. As the V3Fusion-LED model is trained, it learns to generate the correct token sequence by utilizing the information provided by each candidate answer and the V3Fusion-LED evaluation results.

\section{Efficiency Analysis} 

\begin{table}[hbt!]
    \captionof{table}{Inference time and allocated/consumed memory comparison with single VLMs in the model pool.}
    \begin{adjustbox}{width=0.6\textwidth, center}
    \centering
    \begin{tabular}{l c c c c }
    \hline
    \multirow{2}{*}{Model Name} & \multirow{2}{*}{ID} & Inf. & \multicolumn{2}{c}{Mem. (MB)} \\
    \cline{3-5}
    &  & Mean(ms) & Alloc. & Cons.  \\
    \hline
    llava-v1.6-Vic.-7b & 0 & 350.71 & 13472.43 & 7.43 \\
    llava-v1.6-Vic.-13b & 1 & 511.25 & 25587.72 & 7.26 \\
    Qwen2.5-VL-7B-Ins. & 2 & 551.48 & 31774.13 & 4.02 \\
    InternVL2-8B & 3 & 558.66 & 15917.26 & 1.08 \\
    deepseek-vl2-tiny & 4 & 206.14 & 6548.17 & 0.08 \\
    deepseek-vl2-small& 5 & 727.02 & 30928.16 & 0.06 \\
    \hline
    V3Fusion & 234 & 768.02 & 54239.56 & 5.18 \\
    \hline
    \end{tabular}
    \end{adjustbox}
    \label{tab:base_models}
\end{table}

\begin{table}[hbt!]
    \captionof{table}{The effect of training data dependency on VL3Fusion for OCR-VQA, MMMU, and A-OKVQA datasets.
    }
    \begin{adjustbox}{width=0.6\textwidth, center}
    \centering
        \begin{tabular}{c c c c c c}
            \hline
            \multirow{2}{1.4cm}{\centering Train Data (\%) } & \multicolumn{3}{c}{OCR-VQA} & MMMU & A-OKVQA \\
            \cline{2-6}
            &  BLUE-1 & EM & F1 & Acc (\%) & Acc (\%)\\
            \hline
            1 & 15.70 & 0.03 & 26.84 & 42.41 & 87.69 \\
            5 & 82.32 & 68.58 & 83.34 & 52.33 & 87.78 \\
            10 & 84.53 & 71.01 & 85.26 & 52.17 & 87.70 \\
            30 & 85.57 & 71.51 & 86.29 & 52.84 & 87.26 \\
            50 & 86.13 & 72.34 & 86.76 & 52.84 & 86.99 \\
            75 & 86.12 & 71.84 & 86.74 & 54.22 & 86.48 \\
            100 & 86.27 & 71.78 & 86.86 & 54.50 & 88.31 \\
            \hline
        \end{tabular}
        \end{adjustbox}
    \label{tab:train_perc}
\end{table}

Table~\ref{tab:base_models} reports V3Fusion in total inference cost (time, memory) against individual VLMs in our model pool. The Mean inference time for 1000 samples of OCR-VQA is given. V3Fusion loads the pruned model pool in parallel, and the reflection phase takes 41ms. Compared to the top-performing single VLM (Qwen2.5-VL-7B), V3Fusion achieves the best overall performance at a reasonable cost, with an inference time of 768.02ms (Qwen2.5: 551.48ms) and 5.18 MB of memory consumed (Qwen2.5: 4.02 MB). In addition, our training data dependency is lightweight. Table~\ref{tab:train_perc} provides the effectiveness of training data in varying fractions. It shows that using only 5\% of data for training V3Fusion to perform fusion is effective, achieving as good performance as using a large fraction (50~100\%).

\section{Correlation Analysis on Focal Metrics}

\begin{figure}[hbt!]
    \centering
    \includegraphics[width=0.7\linewidth]{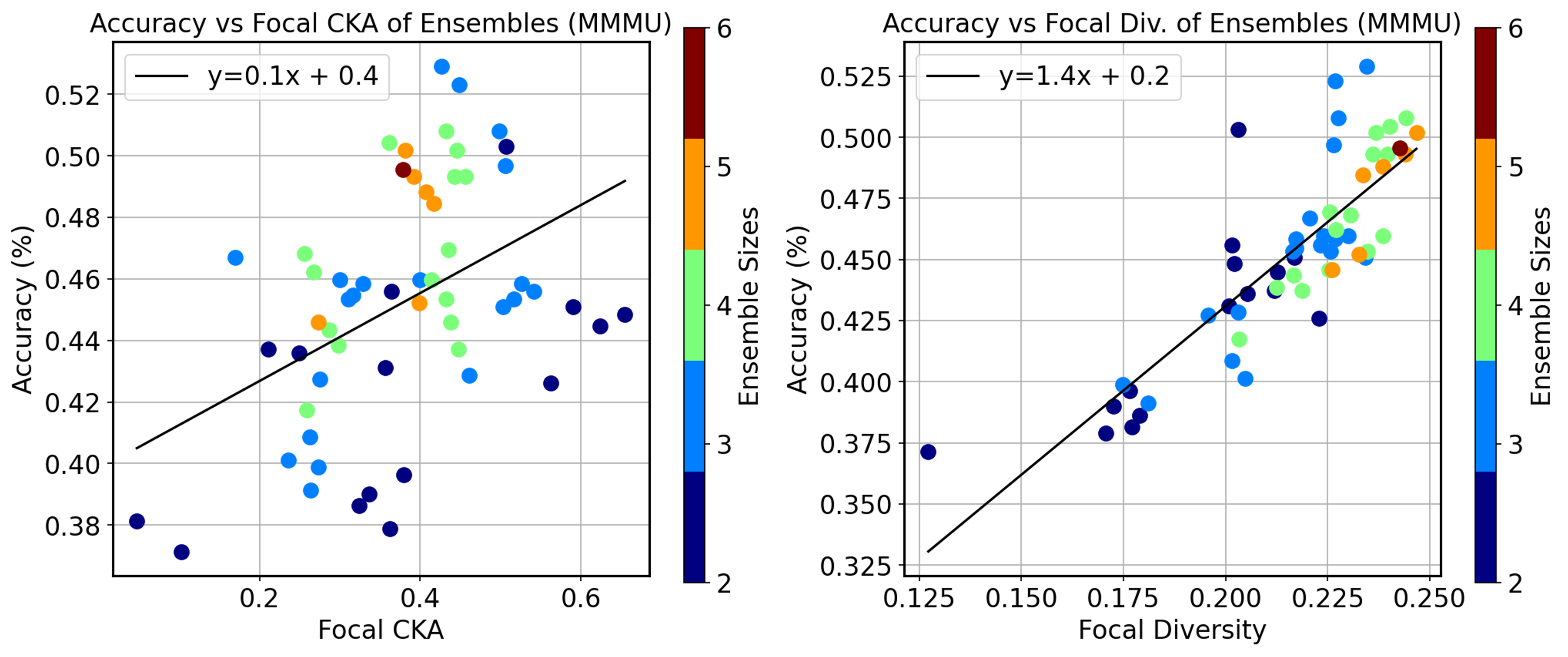}
    \caption{The Correlation between the Accuracy of the ensemble combinations with different sizes and the Focal-CKA on left, and Focal-Diversity on Right.}
    \label{fig:correlation}
\end{figure}

In section \ref{sec:ensemble_pruning}, we presented the ensemble pruning metrics to allow us to select the best ensemble set in terms of performance, which is measured by accuracy. A more direct analysis is to compare performance against diversity. The diversity we aim for has two dimensions: visual perception and language semantics. We capture the first with Focal-CKA with relative dissimilarity among perceptual understanding in an ensemble set. The second is captured by the Focal-diversity, focusing on the output generated by the language decoder. Therefore, in Figure \ref{fig:correlation}, we plot all the possible ensemble set combinations with different sizes and measure their Accuracy and Focal metrics when they team up. In both metrics, as shown in Figure \ref{fig:correlation}, the performance is positively correlated with the diversity metrics. The best-fit line with a positive slope shows the degree of correlation. Overall, the Figure \ref{fig:scalibility} is the surface created by both metrics, and the pruning algorithm (boosted by GA) finds the best combination on that surface.

\section{Finetuning Base Models}

\begin{table}[hbt!]
\captionof{table}{SFT best performing base models vs V3Fusion in OCR-VQA and MMMU datasets.}
    \begin{adjustbox}{width=0.8\textwidth, center}
    \centering
    \begin{tabular}{l c c c c c}
        \hline
        \multirow{2}{1.4cm}{\centering Train Data (\%)} & \multirow{2}{1.2cm}{\centering Train. Time(h)} & \multicolumn{3}{c}{OCR-VQA} & MMMU \\
        \cline{3-6}
        & & BLUE-1 & EM & F1 & Acc (\%)\\
        \hline
        llava-v1.6-vicuna-7b-hf & 4.40 & 50.27 & 30 & 53.88 & 29.06 \\
        llava-v1.6-vicuna-13b-hf & 7.02 & 60.23 & 44 & 63.98 & 30.55 \\
        Qwen2.5-VL-7B-Instruct & 2.17 & 85.12 & \textbf{74.89} & 86.5 & 27.95 \\
        \hline
        V3Fusion &  1.39 & \textbf{86.24} & 71.91 & \textbf{86.82} & \textbf{56.09} \\
        \hline
    \end{tabular}
    \end{adjustbox}
    \label{tab:sft}
\end{table}

A more direct way to measure the gain by ensemble learning is to compare with the individual model performances when they are finetuned and evaluated with the same datasets. The base models, however, have a significant amount of parameters compared to V3Fusion's summary model (LED) and MCQ model (MLP). The training time, therefore, is $7\times$ longer than V3Fusion as reflected in Table \ref{tab:sft}. V3Fusion improves in both datasets, OCR-VQA and MMMU, with the highest F1 and Accuracy scores and lower training time. Especially in the MMMU task, V3Fusion shows $25.54\%$ higher accuracy than the best performing finetuned base model.

\end{document}